\documentclass[twoside,11pt]{article}

\usepackage{jmlr2e}
\usepackage{subcaption}
\usepackage{caption}

\jmlrheading{1}{}{}{}{}{}

\firstpageno{1}

\begin{document}

\title{Long-run Behaviour of Multi-fidelity Bayesian Optimisation}

\author{\name Gbetondji J-S Dovonon \email gbetondji@matterhorn.studio \\
       \addr University College London\\
       Matterhorn Studio\\
       \AND
       \name Jakob Zeitler \email jakob@matterhorn.studio \\
       \addr Matterhorn Studio}


\maketitle

\begin{abstract}
Multi-fidelity Bayesian Optimisation (MFBO) has been shown to generally converge faster than single-fidelity Bayesian Optimisation (SFBO) (\cite{poloczek2017multi}). Inspired by recent benchmark papers, we are investigating the long-run behaviour of MFBO, based on observations in the literature that it might under-perform in certain scenarios (\cite{mikkola2023multi}, \cite{eggensperger2021hpobench}). An under-performance of MBFO in the long-run could significantly undermine its application to many research tasks, especially when we are not able to identify when the under-performance begins. We create a simple benchmark study, showcase empirical results and discuss scenarios and possible reasons of under-performance.

\end{abstract}


\section{Introduction}

The optimisation of costly-to-evaluate functions is a significant challenge in hyperparameter-optimisation in machine learning, material science, drug discovery and more (\cite{kandasamy2020tuning}, \cite{liang2021benchmarking}, \cite{bellamy2022batched}). Bayesian Optimisation (BO, \cite{frazier2018tutorial}) has become an established method to tackle these challenges, also due to the fact that it does not require access to a gradient. As part of so called \textit{grey-box} BO methods (\cite{astudillo2021thinking}) that take into account internal structure of an optimisation problem, \textit{multi-fidelity} BO (MBFO) has emerged as popular method to utilise access to different information sources on the same optimisation problem (\cite{huang2006sequential}).

\section{Review of MFBO}

The core assumption of MFBO is that we have access to different auxiliary fidelities that inform us on our target fidelity. Further assumptions on the nature of those auxiliary fidelities and their relationship to the target fidelity will determine the type of algorithm to use. 
\cite{poloczek2016multiinformation} provide one of the earliest implementations of the MFBO process which has established itself as the predominant choice. They adapt the knowledge gradient acquisition function to the MFBO setting, and provide a new set of mean and covariate functions. Two noteworthy design choices here are the continuous fidelities and the use of a multi-output gaussian process to model both the objective and fidelity. Similar extensions have been made for maximum entropy search (\cite{takeno2020multifidelity}) and expected improvement (\cite{irshad2023leveraging,daulton2020differentiable}) acquisition functions.
While several methods come with attractive theoretical guarantees, when applied to domains where Bayesian optimization is the preferred choice, it is common to see heuristics. On benchmarks like HPOBench (\cite{eggensperger2021hpobench}), heuristic-driven methods dominate (\cite{awad2021dehb,cowenrivers2022hebo}).



\section{Problem and Solution}

Multi-fidelity BO (MFBO) is generally accepted to outperform single-fidelity BO approaches (SFBO). Only recently, the literature has focused on possible failure-modes of MFBO, that could rank it below established SFBO performance. This is partially driven by a range of new and more reliable benchmarks that allow fair comparison of BO algorithms, such as the HPOBench (\cite{eggensperger2021hpobench}) that utilise execution containers to standardise comparison across compute environments. 

\begin{figure}
    \centering
    \includegraphics[width=0.95\textwidth]{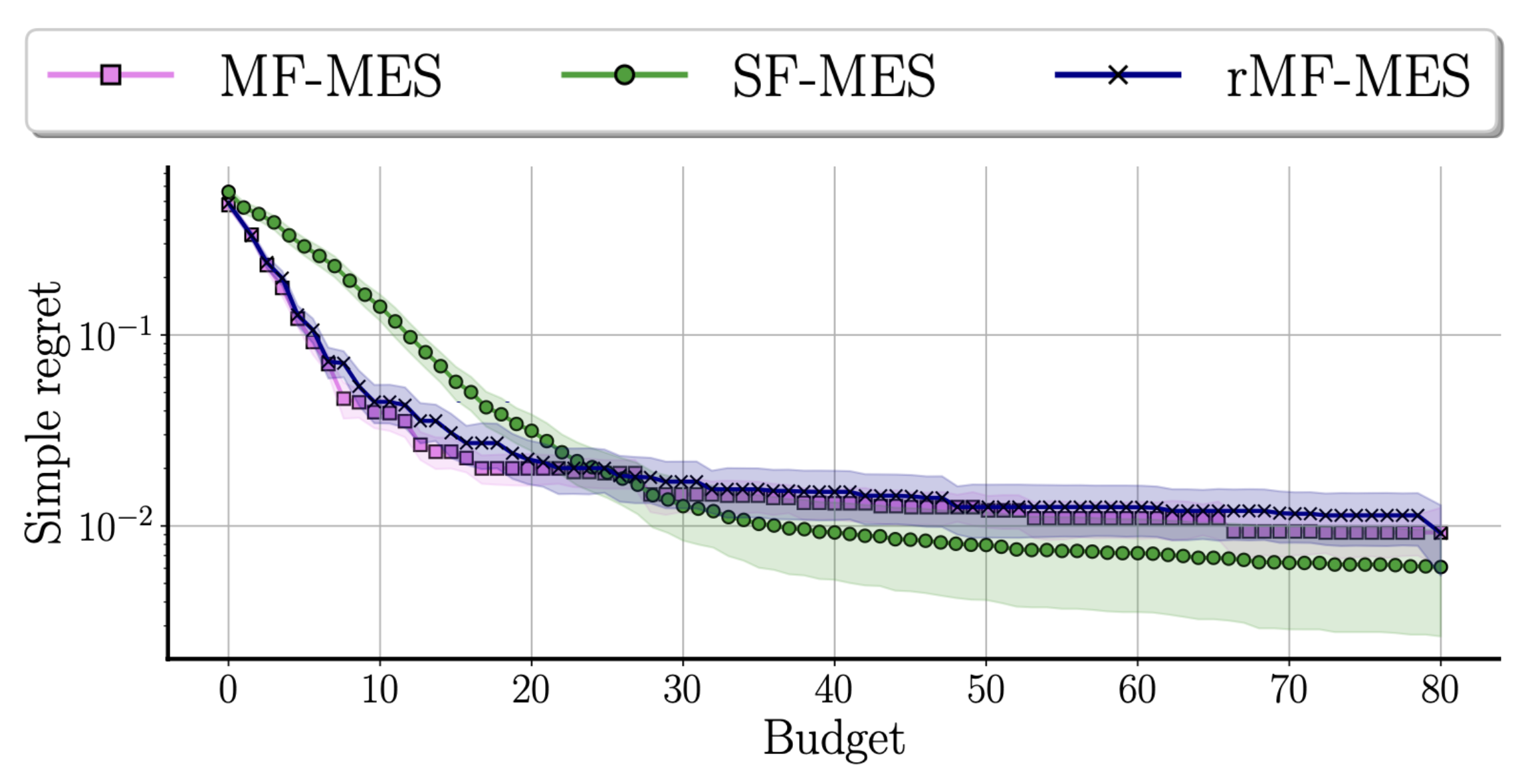}
    \caption{Possible long-run underperformance of MFBO as found in (\cite{mikkola2023multi}, Figure 1). The plot shows maximum-entropy search for single-fidelity (green, SF-MES), multi-fidelity (pink, MF-MES) and robust multi-fidelity (purple, rMF-MES, the method suggested by (\cite{mikkola2023multi}) to handle unreliable information sources).}
    \label{fig:unreliable_information}
\end{figure}

(\cite{mikkola2023multi}) recently evaluated the impact of unreliable information sources on MFBO performance. As part of their investigation, they compared MFBO and SFBO on the Hartmann6D test function, see Figure \ref{fig:unreliable_information}. We can observe a \textit{cross-over point} at a budget of about 25, where the SF-MES starts to outperform the multi-fidelity methods. The confidence intervals still overlap up to the final budget point of 80 and as such do not allow us to conclude that single-fidelity, on average, is outperforming multi-fidelity approaches in the long-run.

\begin{figure}
    \centering
    \includegraphics[width=0.85\textwidth]{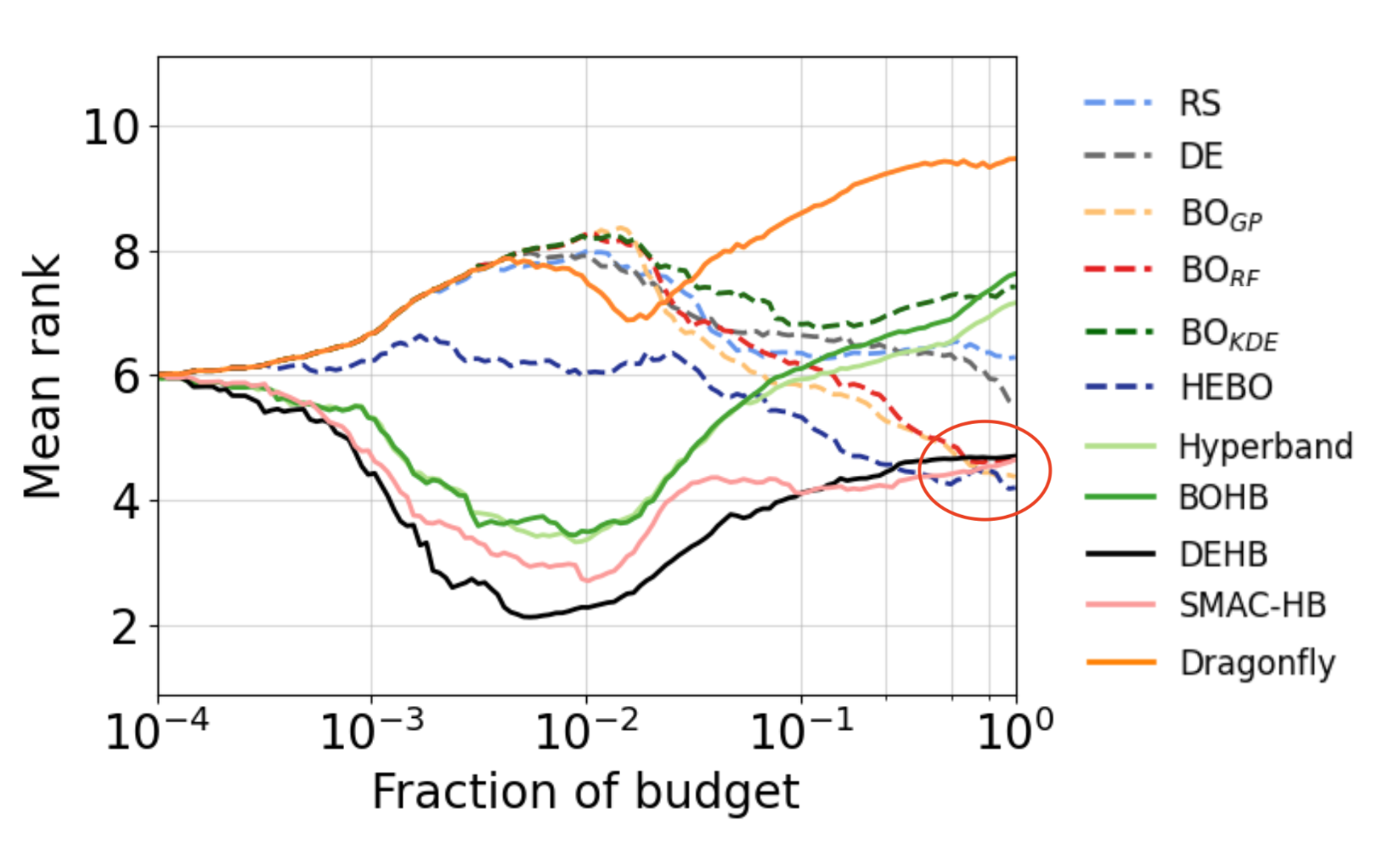}
    \caption{Possible long-run underperformance of MFBO as found in (\cite{eggensperger2021hpobench}, Figure 4). The plot shows the \textit{mean rank} of single-fidelity (dotted) and multi-fidelity (not dotted) methods over increasing \textit{fraction of budget spent}.}
    \label{fig:hpo_bench}
\end{figure}

Evaluation plots created with the HPOBench provide an additional perspective on our observation of a cross-over point in \ref{fig:unreliable_information}. (\cite{eggensperger2021hpobench}) show in Figure \ref{fig:hpo_bench} the mean rank of single-fidelity (dotted) and multi-fidelity (not dotted) methods over increasing fraction of budget spent. While the multi-fidelity methods clearly rank higher for most of the budget spend, we can observe a long-term trend of single-fidelity methods steadily improving in rank and outperforming the other methods at the very right edge of the plot (marked by a red circle). 
The nature of the budget used in HPOBench, a finite set of training data points, does not allow increasing the budget unless the dataset itself is increased. We hypothesize that in settings where hyperparameter optimization can be done efficiently on much larger datasets that single-fidelity might continue to outperform, in line with our observations in Figure \ref{fig:unreliable_information}. 

\begin{figure}
    \centering
    \subcaptionbox{Simple regret vs budget Hartmann6D using MES}{
    \includegraphics[width=0.99\textwidth]{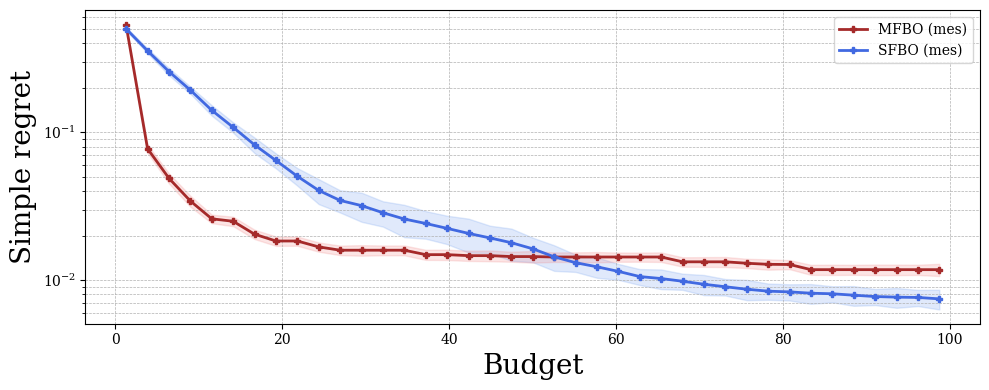}}
    \subcaptionbox{Simple regret vs budget Hartmann6D using KG}{
    \includegraphics[width=0.99\textwidth]{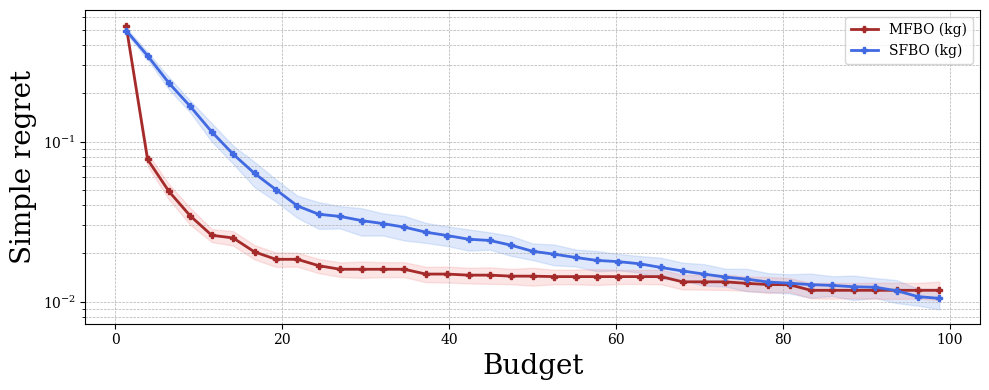}}
    \subcaptionbox{Simple regret vs budget XGB using MES}{
    \includegraphics[width=0.99\textwidth]{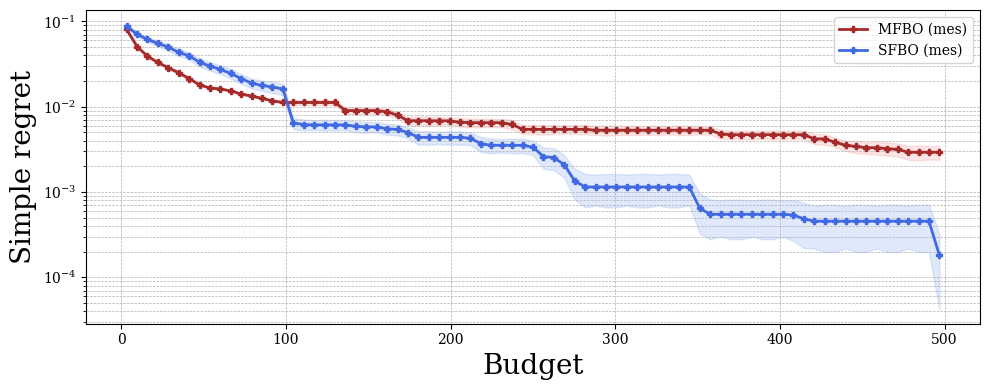}}
    \caption{Our results of 100 trails each for SFBO and MFBO, at budget 100 for Hartmann6D and 500 for XGB, plotted with log-transformation on the y-axis. On both benchmarks, SFBO eventually outperforms MFBO. The crossing point is different with SFBO overtaking MFBO around a budget of 50 on Hartmann6D but 100 on XGB when using MES.}
    \label{fig:hartmann6dcomparison}
\end{figure}

To study our observations in Figure \ref{fig:unreliable_information} and Figure \ref{fig:hpo_bench}, we devise our own simulation runs based on the BoTorch Hartmann6D tutorial where the Hartmann6D test function is augmented with a continuous fidelity choice (\cite{botorch_tutorial}). Our results are shown in Figure \ref{fig:hartmann6dcomparison}. Plotting MFBO and SFBO in a single plot requires adjusting MFBO to the equidistant budget query points of SFBO. While SFBO with a query cost of 1 will query 100 samples with a budget of 100, MFBO will query more than 100 samples (i.e. low and high fidelity), most of the time not at the same budget point as SFBO. Hence, we 'normalise' our MFBO results by querying and reporting the high fidelity every time the MFBO run budget crosses over a SFBO budget point. This does not perfectly represent the MFBO decision making, but aligns the calculation of confidence intervals which leads to a more fair and interpretable comparison plot.

\section{Discussion}

We are able to reproduce the observations made in previous papers, as discussed in Figure \ref{fig:unreliable_information} and \ref{fig:hpo_bench}. We observe the majority of low-fidelity queries occur at the very beginning of the run. MFBO seems to build a 'warm-start' set of low-fidelity observations which then informs a further run dominated by high-fidelity queries. That is why we see such strong performance of MFBO compared to SFBO in the initial budget range of 0 to 15, which is also usually observered in the MFBO literature. 

Our initial results suggest both theoretical and practical next steps. It is important to collect more empirical results on different test functions and BO surrogates and acquisition functions, as our Figures only represent a small set of BO setups such that we cannot conclude that long-run underperformance of MFBO also occurs in other MFBO settings. More importantly, theoretical investigation needs to uncover the reason why we observere MFBO underperformance in Figure \ref{fig:hartmann6dcomparison}.
We consider the following scenarios:

\begin{itemize}
    \item \textbf{Lack of standardisation:} For a fair comparison, we will need to establish agreed standards and implementation strategies. HPOBench (\cite{eggensperger2021hpobench}) attempts to establish such standard and partially succeed, also showcasing the significant impact a lack of standardisation can have. In Figure 4 in their paper, the upper row shows the non-standardise benchmark run, where single-fidelity methods have no clear trend. The bottom row instead, standardised to the benchmark, shows a clear trend of single-fidelity methods possibly outperforming multi-fidelity methods if given enough budget (i.e. the central question of our paper).
    \item \textbf{Application variety:} The problem structure, whether real-world or test function, presumably has an impact on MFBO performance. Indeed, (\cite{mikkola2023multi}) study that exact question and in their Figure 1 showcase how unreliable fidelities render MFBO inferior to SFBO. Our empirical study, although trying to replicate their study of informative auxiliary information sources, might not be the best representation of problem spaces that are challenging to MFBO, and so we hope to expand our study beyond into a wider variety of test functions.
    \item \textbf{No-free-lunch theorem:} Given the significant outperformance of MFBO in the short-term over SFBO, it is reasonable to assume on an intuitive level, that long-term MFBO will suffer from inefficiencies it traded for superior efficiency in the short-term. We hope to explore this idea on a theoretical basis as a next step.
    \item \textbf{Compounding errors:} Considering that lower fidelities are "noisier" than higher fidelities, it is possible that the errors in measurements when using the lowest fidelities accumulates, leading the optimization process to get stuck in a local minima in the long-run.
\end{itemize}

\section{Conclusion}

We studied long-run behaviour of multi-fidelity Bayesian Optimisation (MBFO), observing in the literature a possible under-performance compared to single-fidelity Bayesian optimisation (SFBO). Our own empirical studies provide further evidence on these observations. With a multitude of MFBO algorithms available \footnote{e.g. see Matterhorn Studio's OptStore for an overview: \url{https://matterhorn.studio}}, it is important to evaluate their limitations for the best outcome in applications of adaptive experimentation. We discussed a few possible scenarios and hope to expand our empirical studies to characterise the long-run behaviour of MFBO for a variety of applications beyond the test functions studied so far.


\acks{Thank you to the workshop reviewers for their detailed feedback that allowed us to improve and clarify our work. We are also grateful for the support for this work from the UKRI Innovate UK Transformative Technologies Grant 2023 Series.}


\vskip 0.2in
\bibliography{sample}

\begin{thebibliography}{16}
\providecommand{\natexlab}[1]{#1}
\providecommand{\url}[1]{\texttt{#1}}
\expandafter\ifx\csname urlstyle\endcsname\relax
  \providecommand{\doi}[1]{doi: #1}\else
  \providecommand{\doi}{doi: \begingroup \urlstyle{rm}\Url}\fi

\bibitem[Astudillo and Frazier(2021)]{astudillo2021thinking}
Raul Astudillo and Peter~I Frazier.
\newblock Thinking inside the box: A tutorial on grey-box bayesian optimization.
\newblock In \emph{2021 Winter Simulation Conference (WSC)}, pages 1--15. IEEE, 2021.

\bibitem[Awad et~al.(2021)Awad, Mallik, and Hutter]{awad2021dehb}
Noor Awad, Neeratyoy Mallik, and Frank Hutter.
\newblock Dehb: Evolutionary hyperband for scalable, robust and efficient hyperparameter optimization, 2021.

\bibitem[Balandat(2021)]{botorch_tutorial}
Balandat.
\newblock Continuous multi-fidelity bo in botorch with knowledge gradient, 2021.
\newblock URL \url{https://botorch.org/tutorials/multi_fidelity_bo}.

\bibitem[Bellamy et~al.(2022)Bellamy, Rehim, Orhobor, and King]{bellamy2022batched}
Hugo Bellamy, Abbi~Abdel Rehim, Oghenejokpeme~I Orhobor, and Ross King.
\newblock Batched bayesian optimization for drug design in noisy environments.
\newblock \emph{Journal of Chemical Information and Modeling}, 62\penalty0 (17):\penalty0 3970--3981, 2022.

\bibitem[Cowen-Rivers et~al.(2022)Cowen-Rivers, Lyu, Tutunov, Wang, Grosnit, Griffiths, Maraval, Jianye, Wang, Peters, and Ammar]{cowenrivers2022hebo}
Alexander~I. Cowen-Rivers, Wenlong Lyu, Rasul Tutunov, Zhi Wang, Antoine Grosnit, Ryan~Rhys Griffiths, Alexandre~Max Maraval, Hao Jianye, Jun Wang, Jan Peters, and Haitham~Bou Ammar.
\newblock Hebo pushing the limits of sample-efficient hyperparameter optimisation, 2022.

\bibitem[Daulton et~al.(2020)Daulton, Balandat, and Bakshy]{daulton2020differentiable}
Samuel Daulton, Maximilian Balandat, and Eytan Bakshy.
\newblock Differentiable expected hypervolume improvement for parallel multi-objective bayesian optimization, 2020.

\bibitem[Eggensperger et~al.(2021)Eggensperger, M{\"u}ller, Mallik, Feurer, Sass, Klein, Awad, Lindauer, and Hutter]{eggensperger2021hpobench}
Katharina Eggensperger, Philipp M{\"u}ller, Neeratyoy Mallik, Matthias Feurer, Ren{\'e} Sass, Aaron Klein, Noor Awad, Marius Lindauer, and Frank Hutter.
\newblock Hpobench: A collection of reproducible multi-fidelity benchmark problems for hpo.
\newblock \emph{arXiv preprint arXiv:2109.06716}, 2021.

\bibitem[Frazier(2018)]{frazier2018tutorial}
Peter~I. Frazier.
\newblock A tutorial on bayesian optimization, 2018.

\bibitem[Huang et~al.(2006)Huang, Allen, Notz, and Miller]{huang2006sequential}
Deng Huang, Theodore~T Allen, William~I Notz, and R~Allen Miller.
\newblock Sequential kriging optimization using multiple-fidelity evaluations.
\newblock \emph{Structural and Multidisciplinary Optimization}, 32:\penalty0 369--382, 2006.

\bibitem[Irshad et~al.(2023)Irshad, Karsch, and Döpp]{irshad2023leveraging}
Faran Irshad, Stefan Karsch, and Andreas Döpp.
\newblock Leveraging trust for joint multi-objective and multi-fidelity optimization, 2023.

\bibitem[Kandasamy et~al.(2020)Kandasamy, Vysyaraju, Neiswanger, Paria, Collins, Schneider, Poczos, and Xing]{kandasamy2020tuning}
Kirthevasan Kandasamy, Karun~Raju Vysyaraju, Willie Neiswanger, Biswajit Paria, Christopher~R Collins, Jeff Schneider, Barnabas Poczos, and Eric~P Xing.
\newblock Tuning hyperparameters without grad students: Scalable and robust bayesian optimisation with dragonfly.
\newblock \emph{The Journal of Machine Learning Research}, 21\penalty0 (1):\penalty0 3098--3124, 2020.

\bibitem[Liang et~al.(2021)Liang, Gongora, Ren, Tiihonen, Liu, Sun, Deneault, Bash, Mekki-Berrada, Khan, et~al.]{liang2021benchmarking}
Qiaohao Liang, Aldair~E Gongora, Zekun Ren, Armi Tiihonen, Zhe Liu, Shijing Sun, James~R Deneault, Daniil Bash, Flore Mekki-Berrada, Saif~A Khan, et~al.
\newblock Benchmarking the performance of bayesian optimization across multiple experimental materials science domains.
\newblock \emph{npj Computational Materials}, 7\penalty0 (1):\penalty0 188, 2021.

\bibitem[Mikkola et~al.(2023)Mikkola, Martinelli, Filstroff, and Kaski]{mikkola2023multi}
Petrus Mikkola, Julien Martinelli, Louis Filstroff, and Samuel Kaski.
\newblock Multi-fidelity bayesian optimization with unreliable information sources.
\newblock In \emph{International Conference on Artificial Intelligence and Statistics}, pages 7425--7454. PMLR, 2023.

\bibitem[Poloczek et~al.(2016)Poloczek, Wang, and Frazier]{poloczek2016multiinformation}
Matthias Poloczek, Jialei Wang, and Peter~I. Frazier.
\newblock Multi-information source optimization, 2016.

\bibitem[Poloczek et~al.(2017)Poloczek, Wang, and Frazier]{poloczek2017multi}
Matthias Poloczek, Jialei Wang, and Peter Frazier.
\newblock Multi-information source optimization.
\newblock \emph{Advances in neural information processing systems}, 30, 2017.

\bibitem[Takeno et~al.(2020)Takeno, Fukuoka, Tsukada, Koyama, Shiga, Takeuchi, and Karasuyama]{takeno2020multifidelity}
Shion Takeno, Hitoshi Fukuoka, Yuhki Tsukada, Toshiyuki Koyama, Motoki Shiga, Ichiro Takeuchi, and Masayuki Karasuyama.
\newblock Multi-fidelity bayesian optimization with max-value entropy search and its parallelization, 2020.

\end{thebibliography}

\end{document}